\documentclass[10pt,twocolumn,letterpaper]{article}

\usepackage{cvpr}              % To produce the CAMERA-READY version

\usepackage{url}
\usepackage{graphicx}
\usepackage{paralist}
\usepackage{bm}
\usepackage{amsmath}
\usepackage{multirow}
\usepackage{xr}
\usepackage{booktabs}
\usepackage[leftcaption]{sidecap}
\usepackage{booktabs} % For formal tables
\usepackage{sidecap}
\usepackage{subcaption}
\usepackage{stfloats}
\usepackage{float}

\usepackage[dvipsnames]{xcolor}

\definecolor{cvprblue}{rgb}{0.21,0.49,0.74}
\usepackage[pagebackref,breaklinks,colorlinks,allcolors=cvprblue]{hyperref}

\def\paperID{579} % *** Enter the Paper ID here
\def\confName{CVPR}
\def\confYear{2025}

\title{MikuDance: Animating Character Art with Mixed Motion Dynamics}

\author{
    Jiaxu Zhang$^{1, 2}$\qquad
    Xianfang Zeng$^{2}$\footnotemark[2]\qquad
    Xin Chen$^{3}$\qquad 
    Wei Zuo$^{2}$\qquad 
    Gang Yu$^{2}$\footnotemark[3] \qquad
    Zhigang Tu$^{1}$\footnotemark[3]\\
    $^{1}$Wuhan University\qquad
    $^{2}$StepFun \qquad
    $^{3}$ByteDance \qquad \\
    {\tt \small Project page: \url{https://kebii.github.io/MikuDance}}
    }

\begin{document}

% \maketitle

\twocolumn[{%
    \centering
    \vspace{-.8cm}
    \includegraphics[width=2cm]{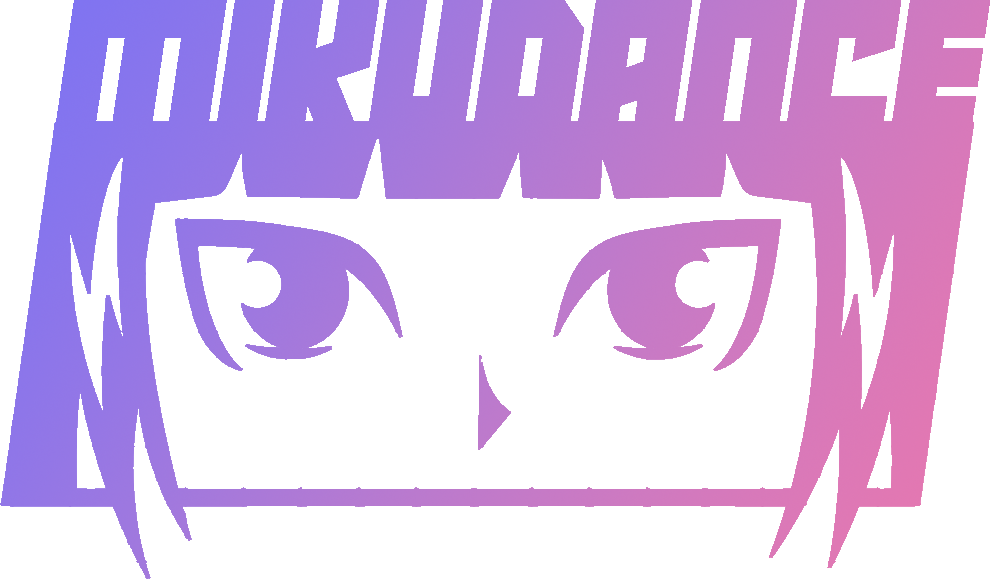}\\[\baselineskip] % 调整宽度控制logo大小
    \vspace{-1.5cm} % 控制logo和标题之间的间距
    \maketitle
}]

\renewcommand{\thefootnote}{\fnsymbol{footnote}}
% \footnotetext[1]{Most of this work was done during Jiaxu’s internship at StepFun.}
\footnotetext[2]{Xianfang Zeng is the project leader.}
\footnotetext[3]{Corresponding authors: skicy@outlook.com, tuzhigang@whu.edu.cn}
\renewcommand{\thefootnote}{\arabic{footnote}}

% \maketitle

\begin{abstract}
We propose MikuDance, a diffusion-based pipeline incorporating mixed motion dynamics to animate stylized character art.
MikuDance consists of two key techniques: Mixed Motion Modeling and Mixed-Control Diffusion, to address the challenges of high-dynamic motion and reference-guidance misalignment in character art animation.
Specifically, a Scene Motion Tracking strategy is presented to explicitly model the dynamic camera in pixel-wise space, enabling unified character-scene motion modeling. Building on this, the Mixed-Control Diffusion implicitly aligns the scale and body shape of diverse characters with motion guidance, allowing flexible control of local character motion. Subsequently, a Motion-Adaptive Normalization module is incorporated to effectively inject global scene motion, paving the way for comprehensive character art animation.
Through extensive experiments, we demonstrate the effectiveness and generalizability of MikuDance across various character art and motion guidance, consistently producing high-quality animations with remarkable motion dynamics.
\end{abstract}    
\section{Introduction}

Character art plays a crucial role in the film, game, and digital design industries. Animating character art, which brings static character images to life, has been an increasingly prominent challenge in computer vision and graphics. Traditional animation software, such as MMD \cite{MikuMikuDance}, and Live2D \cite{Live2D}, requires professional skills, posing significant barriers for non-experts. Recently, image-to-video generation methods \cite{wu2023tune, ma2024follow, guo2024liveportrait, li2024generative, zhu2024poseanimate} have emerged as a promising solution for animation. However, these methods are primarily designed for animating real-world humans and cannot be directly applied to character art due to the following two key challenges.

\begin{figure}[t]
% \vspace{-.4cm}
\setlength{\abovecaptionskip}{-.2cm}
\setlength{\belowcaptionskip}{-.5cm}
\begin{center}
   \includegraphics[width=1.0\linewidth]{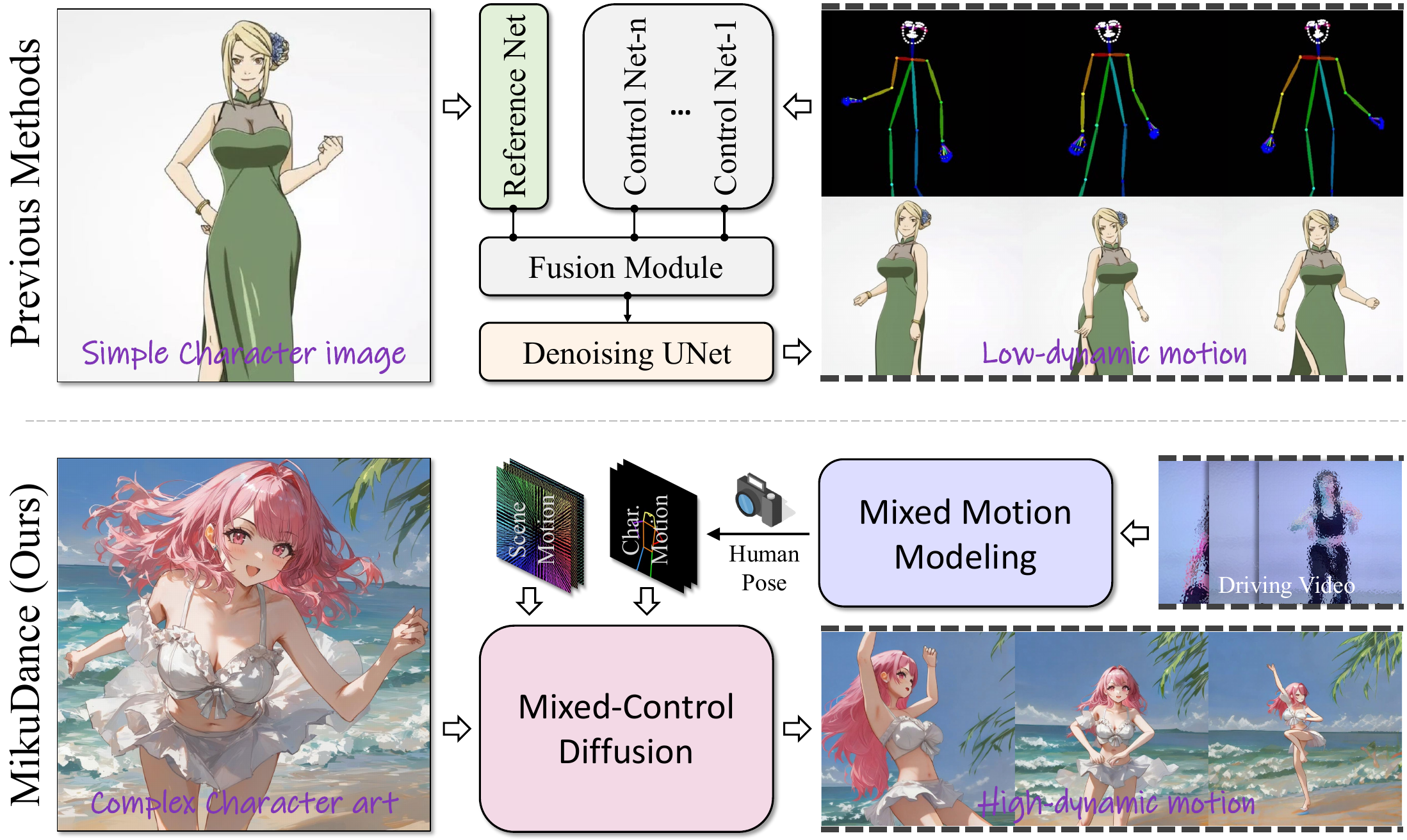}
\end{center}
   \caption{\textbf{We propose MikuDance}, a Diffusion-based pipeline for animating complex and stylized character art with high-dynamic motion guidance. The core insight of MikuDance lies in its \textit{Mixed Motion Modeling} and \textit{Mixed-Control Diffusion} capabilities.}
\label{fig:1}
\end{figure}

%
% The first challenge arises from the complex interplay between the foreground and background in character art, making it difficult to control large-scale motions for both elements during animation.
%
The first challenge arises from the \textit{high-dynamic motion guidance} for both the complex foreground and background in character art, making unified control and maintaining temporal consistency difficult. For instance, in the second drawing shown in Figure \ref{fig:1}, the girl is portrayed in an elegant dress against an artistic background, driven by large-scale dance motion and camera movements. Existing image animation methods, such as Animate Anyone \cite{hu2024animate} and DISCO \cite{wang2024disco}, are primarily limited to animating humans with a static camera and a clean background. In contrast, animating character art requires the model to handle large-scale motion within complex scenes. As a result, simultaneously modeling the high-dynamic motion of both characters and the entire backgrounds becomes a critical task.

The second challenge stems from the \textit{unique body shapes and diverse scales} of characters, which often misalign with motion guidance. For example, anime characters exhibit a large head-to-body ratio, exaggerated poses, and varied artistic styles. As shown in Figure \ref{fig:1}, previous methods employ separate networks to process the reference image and motion guidance, oversimplifying the task by assuming a pre-aligned human body \cite{zhu2024champ, shao2024human4dit}, or performing alignment through pre-processing \cite{wang2024unianimate}, which often leads to unnecessarily complex motion control architectures. However, considering that art images feature distinct characters, explicit alignment becomes impractical. Thus, implicitly aligning the reference image and motion guidance within a unified structure presents a significant task.

To address these challenges and leverage recent advancements in video generation for character art animation, we propose MikuDance. MikuDance animates in-the-wild character art by utilizing mixed character-scene motion guidance to generate videos with large-scale motion dynamics. It introduces two key techniques: Mixed Motion Modeling and Mixed-Control Diffusion.

Mixed Motion Modeling explicitly represents character motions and 3D dynamic camera movements within a unified 2D space, enabling local and global motion guidance of both foreground and background in animations. Unlike previous methods that use camera parameters directly as the control signal \cite{wang2024humanvid, xu2024camco, he2024cameractrl}, we propose a Scene Motion Tracking (SMT) strategy to model the global motion. SMT strategy projects the reference image to a scene point cloud and tracks corresponding points across consecutive camera frames, transforming camera poses into a pixel-wise scene motion representation. This scene motion, resembling keypoint-based character motion, establishes the foundational basis of our mixed motion control approach.

Mixed-Control Diffusion addresses misalignments in scales and body shapes of characters by integrating all reference and motion guidance into a unified Reference UNet \cite{ronneberger2015u}. This design is based on our observation that the mixed and implicit alignment approach outperforms other sophisticated control networks while preserving an elegant model architecture. Moreover, as scene motion guides the global dynamics of the animation, we carefully design a Motion-Adaptive Normalization (MAN) module to flexibly inject the scene motion into the Reference UNet, effectively integrating global dynamics and maintaining local consistency in character art animation.

By leveraging these two techniques and a mixed-source training approach, MikuDance animates diverse character art with mixed motion dynamics. We evaluate MikuDance using a range of reference characters and motion guidance. Both qualitative and quantitative results demonstrate that MikuDance can generate high-quality animation, particularly in maintaining consistency in character local motion and effectively handling large-scale scene motion.

Contributions of our MikuDance are listed below:
\begin{compactitem}
\item Mixed Motion Modeling is proposed to explicitly model character and camera motions within a unified pixel-wise space, enabling the effective representation of high-dynamic motion.
\item Mixed-Control Diffusion is exploited to implicitly align character shape, pose, and scale with the motion guidance, enabling cohesive motion control for character art animation.
\item Extensive experiments demonstrate the effectiveness and generalizability of our MikuDance, achieving superior animation quality and high-dynamic motion control compared to state-of-the-art methods.
\end{compactitem}
\begin{figure*}[t]
\vspace{-.6cm}
\setlength{\abovecaptionskip}{-.2cm}
\setlength{\belowcaptionskip}{-.4cm}
\begin{center}
   \includegraphics[width=1.0\linewidth]{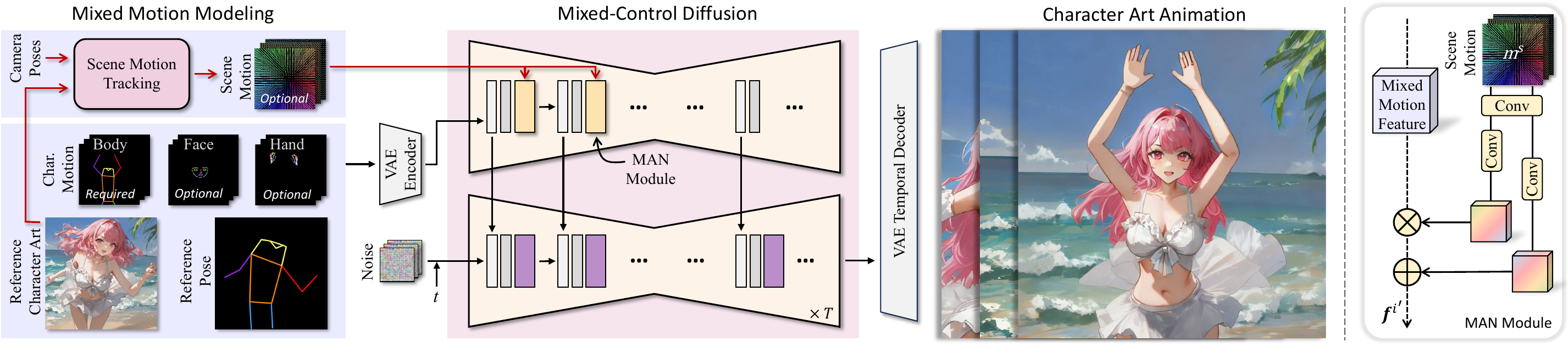}
\end{center}
   \caption{\textbf{Illustration of our MikuDance pipeline.} Given a reference character art and a driving video, the pixel-wise scene motion is predicted using the Scene Motion Tracking (SMT) strategy, which is combined with the character poses to form the character-scene mixed motion guidance. The Mixed-Control Diffusion subsequently generates the animation in a latent space, guided by the character poses and the scene motion injected through the Motion-Adaptive Normalization (MAN) module.}
\label{fig:2}
\end{figure*}

\section{Related Work}
\noindent\textbf{Human image animation}, which aims to generate human action videos from a reference image, has been studied extensively in recent years. Early methods, such as FOMM \cite{siarohin2019first} and Liquid Warping GAN \cite{liu2019liquid}, warped the source human image using affine transformations guided by dense optical flow. Recently, diffusion-based approaches have gained traction due to their strong generalizability \cite{chang2023magicpose}. For instance, PIDM \cite{bhunia2023person} introduced a texture diffusion module to model the correspondence between human appearance and poses. DreamPose \cite{karras2023dreampose} incorporated an adapter module to integrate both CLIP \cite{radford2021learning} and VAE \cite{kingma2013auto} features of reference images. Animate Anyone \cite{hu2024animate} proposed a UNet-based ReferenceNet to extract appearance from reference images, along with a Pose Guider to encode the driving pose sequences, and a temporal module introduced by AnimateDiff \cite{guoanimatediff} to enhance video consistency. 

Following Animate Anyone, many diffusion-based models have been successively proposed, incorporating specific human motion guidance such as dense poses \cite{xu2024magicanimate}, 3D human models \cite{zhu2024champ}, and hand sequences \cite{zhou2024realisdance}. Additionally, DISCO \cite{wang2024disco} proposed decoupling human subjects from backgrounds, and UniAnimate \cite{wang2024unianimate} employed a unified Denoising UNet to generate long-term video.  While these methods introduce increasingly stringent human body priors, they do not account for the variable shapes and scales found in anime characters. Moreover, Animate-X \cite{tan2024animate} generalizes the pipeline to anthropomorphic characters but neglects the broader movements of the entire scene. In contrast, we design the Mixed-Control Diffusion to address the challenges of misalignment and comprehensive motion control in animating character art.

\vspace{1mm}
\noindent\textbf{Controllable video generation} builds upon the success of image generation by integrating additional spatial and temporal control signals. For example, VideoCrafter \cite{chen2023videocrafter1} and DynamiCrafter \cite{xing2023dynamicrafter} control the first frame of the generated video through an image injection module. DragNUWA \cite{yin2023dragnuwa} integrates motion trajectories to control object movements. Animate Anything \cite{dai2023animateanything} introduces area guidance and motion strength guidance to achieve fine-grained motion control. In this work, the central focus is on the unified guidance of both character motion and camera movements.

Moving forward, existing works on camera control in video generation \cite{xu2024camco, he2024cameractrl, yang2024direct} often use Plücker coordinates \cite{jia2020plucker} as an embedding of camera poses. Recent works such as Human4DiT \cite{shao2024human4dit} and HumanVid \cite{wang2024humanvid} consider camera movement in human videos by using an independent camera encoder and directly adding this embedding into the Denoising UNet. However, this camera embedding has a significant domain gap compared to the image pixel-wise guidance of character motion, making it difficult to fuse and maintain consistency in animation. In this work, we explicitly track camera movement in pixel space and integrate it with character motion guidance using the Motion-Adaptive Normalization module, enabling high-dynamic motion modeling in character art animation.
\section{Method}

As illustrated in Figure \ref{fig:2}, given a character art $\mathcal{I}$ and a driving video $\mathcal{V}$, The purpose of our MikuDance is to animate the image $\mathcal{I}$ with reference to the human and camera motion in video $\mathcal{V}$. Specifically, we utilize Xpose \cite{yang2023Xpose} to separately extract pose sequences of the human body, face, and hand, and employ DROID-SLAM \cite{teed2021droid} to extract the camera poses $\{\bm{p}^{c}_l\}^L_{l=1}$, $\bm{p}^{c}\in \mathbb{R}^{L \times 7}$ from $\mathcal{V}$. $L$ indicates the sequence length. The character's initial body pose, which exhibits significant scale and pose differences compared to the driving video, is also extracted from $\mathcal{I}$. Next, the image $\mathcal{I}$ and all the reference and driving pose images are encoded into the latent space through a VAE Encoder. The camera poses $\bm{p}^{c}$ is processed through the Scene Motion Tracking strategy to obtain the pixel-wise scene motion guidance. Then, the Mixed-Control Diffusion is used to animate $\mathcal{I}$ guided by the mixed motion guidance of the character poses and scene motion in the latent space. Finally, the latent output is decoded through the VAE Temporal Decoder to produce the character art animation.

\subsection{Preliminaries on Stable Diffusion} \label{sec:pre}
Stable Diffusion (SD) \cite{rombach2022high} is a popular Latent Diffusion Model for text-to-image generation. SD consists of a VAE \cite{kingma2013auto} for auto-encoding the images, and a UNet \cite{ronneberger2015u} for noise estimation to iteratively transform a noise image into a latent image by the reverse diffusion process \cite{ho2020denoising}. Given an input data distribution $\bm{x}_{0}$, the forward process apply a Markov noising process of $T$ steps on $\bm{x}_{0}$ to obtain $\{\bm{x}_{t}\}^{T}_{t=0}$:
\begin{equation} \label{preEq.1}
q(\bm{x}_{t}|\bm{x}_{t-1}) = \mathcal{N} \left (\sqrt{\alpha_{t}}\bm{x}_{t-1}, (1-\alpha_{t})\bm{I} \right ),
\end{equation}
where $\alpha_{t} \in (0,1)$ are constant hyper-parameters. When $\alpha_{t}$ is small enough, $\bm{x}_{T} \sim \mathcal{N}(\bm{0},\bm{I})$.
The reverse process takes a noisier data distribution $\bm{x}_{t}$ and generates a less noisy distribution $\bm{x}_{t-1}$ using an UNet, which is trained with the simple loss function:
\begin{equation} \label{preEq.2}
\mathcal{L}_{simple} := \mathbb{E}_{\bm{\epsilon}, t, c} \left [\lVert \bm{\epsilon}- \bm{\epsilon}_{\theta}(\bm{x}_{t},t,c)\rVert^{2}_{2} \right ],
\end{equation}
where $\bm{\epsilon}$ is the Gaussian noise. $c$ is the text condition. $\bm{\epsilon}_{\theta}(\cdot)$ is the trainable noise predictor. In this work, inspired by Animate Anyone \cite{hu2024animate}, we utilize the pre-trained SD-1.5 as our base model to develop our animation pipeline, MikuDance.

\begin{figure*}[t]
\vspace{-.6cm}
\setlength{\abovecaptionskip}{-.2cm}
\setlength{\belowcaptionskip}{-.5cm}
\begin{center}
   \includegraphics[width=1.0\linewidth]{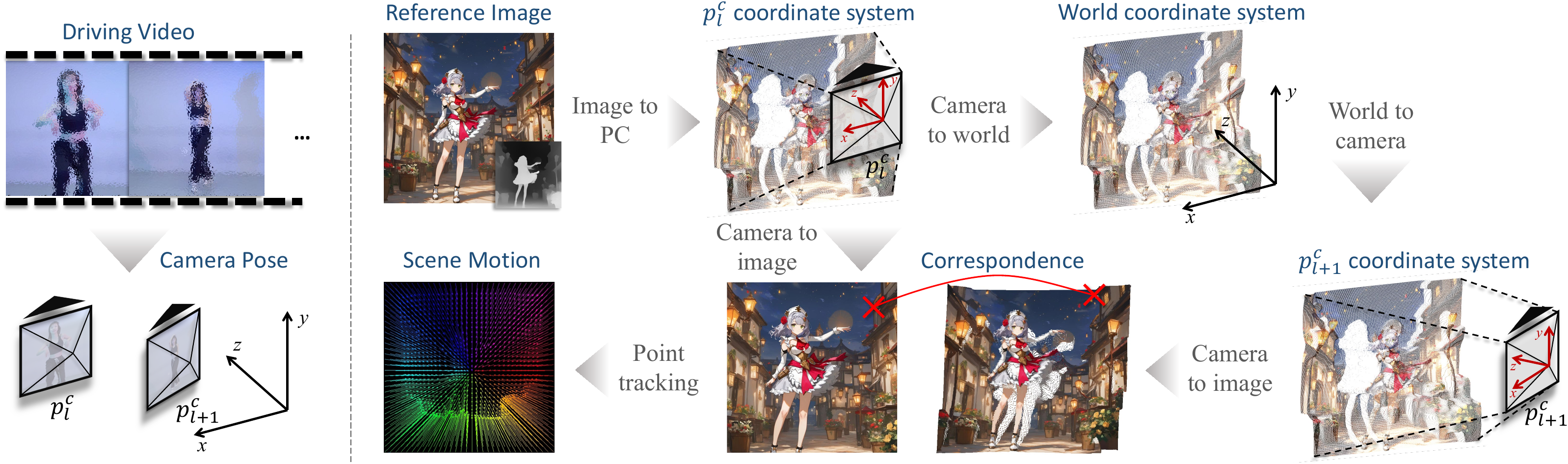}
\end{center}
   \caption{\textbf{Illustration of the Scene Motion Tracking strategy.} To effectively guide global background motion, 3D camera poses extracted from the driving video are transformed into a pixel-wise 2D space through the projection of the scene's point cloud (PC).}
\label{fig:3}
\end{figure*}

\subsection{Mixed Motion Modeling} \label{sec:mmm}
Following existing human image animation methods \cite{hu2024animate, chang2023magicpose, wang2024disco}, we use image-based pose sequences as motion guidance for characters. Unlike previous approaches that directly extract the character's whole-body poses, we separately extract poses for the body, face, and hands, allowing the face and hands to be optional and enabling more flexible motion control. However, character animation often involves high-dynamic motion throughout the entire scene to enhance the visual impact of storytelling. Traditional pose sequences offer only character motion guidance, lacking representations for background dynamics. To address this, we introduce the Scene Motion Tracking strategy.

\vspace{1mm}
\noindent\textbf{Scene Motion Tracking (SMT).} As illustrated in Figure \ref{fig:3}, given a camera pose $\bm{p}^{c}_l$ from the driving video at the $l$-th frame, a scene point cloud $\phi^{l} \in \mathbb{R}^{N \times 3}$ of the character art $\mathcal{I}$ is constructed in the $\bm{p}^{c}_l$ coordinate system using the depth map of $\mathcal{I}$. $N$ is the number of points, which depends on the size of the image. Next, $\phi^{l}$ is transferred into the world coordinate system through the camera-to-world matrix $\mathcal{T}^{l} \in \mathbb{R}^{N \times 4 \times 4}$ of $\bm{p}^{c}_l$, resulting in the point cloud $\phi^{w}$. Subsequently, by applying the world-to-camera matrix $\mathcal{Y}^{l + 1} \in \mathbb{R}^{N \times 4 \times 4}$ of the camera pose $\bm{p}^{c}_{l+1}$, we obtain the point cloud $\phi^{l+1}$ for the next frame. Finally, we project $\phi^{l}$ and $\phi^{l+1}$ into the image coordinate system using the intrinsic matrices $\mathcal{K} \in \mathbb{R}^{N \times 3 \times 4}$ of $\bm{p}^{c}_l$ and $\bm{p}^{c}_{l+1}$, respectively. The projected images are denoted as $\mathcal{I}^{l}$ and $\mathcal{I}^{l+1}$. Since these two projected images are rendered from the same scene point cloud in the world coordinate system, we can calculate the scene motion $\bm{m}^{s} \in \mathbb{R}^{N \times 2}$ in the image space based on the correspondence of the points. This process is formulated as:
\begin{equation} \label{Eq.1}
\left(\bm{z}^{l} - \bm{z}^{l+1} \right) \left[ \begin{matrix} \bm{m}^{s} \\ \bm{1} \end{matrix} \right]  = \mathcal{K}^{l} \left[ \begin{matrix} \phi^{l} \\ \bm{1} \end{matrix} \right] - \mathcal{K}^{l+1} \mathcal{Y}^{l+1} \mathcal{T}^{l} \left[ \begin{matrix} \phi^{l} \\ \bm{1} \end{matrix} \right],
\end{equation}
where $\bm{z}$ represents the projected Z-axis coordinates, serving as a scale factor for the scene motion $\bm{m}^{s}$.

Notably, our SMT strategy diverges from the optical flow commonly used in video generation methods \cite{gengmotion, niu2024mofa} in two key respects: first, the scene motion extracted by SMT is independent of the driving video's content, whereas optical flow is content-dependent. Second, SMT tracks 3D points from the point cloud, while optical flow tracks pixel movements in the image domain, without considering the actual 3D scene. Consequently, our SMT strategy provides decoupled camera dynamic information, which is crucial for continuous background motion in character art animation.

In the proposed SMT process, we assume that the character and scene are static and standardized in the first camera. However, in real applications of character art animation, the reference scene often misaligns with the camera scale of the driving video, and the character pose varies in each frame. This ambiguity cannot be explicitly eliminated and necessitates the model to be implicitly perception guided by the character pose and the art image. Therefore, we propose Mixed-Control Diffusion in the next section.

\subsection{Mixed-Control Diffusion} \label{sec:mcd}
The concept of our Mixed-Control Diffusion is to mix and fuse all motion guidance for the character and scene within a unified reference space, thereby achieving aligned motion control over the animation.

As illustrated in Figure \ref{fig:2}, drawing inspiration from Animate Anyone \cite{hu2024animate}, we utilized the pre-trained SD-1.5 as the base Denoising UNet and a copy of it as the Reference UNet to achieve controllable image-to-video generation. Distinct from Animate Anyone and other related work, we eliminate separate encoders for motion guidance and simultaneously encode the reference character art, the reference pose, and all character pose guidance using the VAE Encoder, embedding them into the same latent space. Next, all embedded guidance is concatenated along the channel dimension to serve as the input for the Mixed-Control Reference UNet. To accommodate this mixed input, we expand the channel of the input convolution layer in the Reference UNet, initializing the added parameters with zero convolution weights \cite{zhang2023adding}. Additionally, the reference image is embedded using the CLIP image encoder \cite{radford2021learning} and serves as the $key$ features in the cross-attention operations of both the Denoising UNet and the Reference UNet. This process is commonly used in existing work and is therefore omitted from Figure \ref{fig:2}.

In each denoising step $t$ of the Mixed-Control Diffusion, the self-attention features from the Reference UNet are injected into the Denoising UNet through an addition operation. The Reference UNet requires inference only once, while the denoising process is repeated $T$ times. The experiments demonstrate that our mixed control approach outperforms other control encoding and fusion methods by effectively addressing the misalignments between the reference image and the motion guidance. Furthermore, since scene motion exerts a global influence on the animation frames, it is intuitive to integrate it with character motion using an adaptive normalization method, as introduced below.

\vspace{1mm}
\noindent\textbf{Motion-Adaptive Normalization (MAN)} is designed to effectively mix the extracted pixel-wise scene motion, $\bm{m}^{s}$, and enhance the temporal consistency of both foreground and background animations.

Inspired by SPADE \cite{park2019semantic}, which employs a spatial-aware normalization method to capture semantics for image synthesis, we propose to implement spatial-aware normalization adapted by scene motion, as shown in the right part of Figure \ref{fig:2}. Let the mixed motion feature of the $i$-th block in the Reference UNet be denoted as $\bm{f}^i$. We first normalize it using the Instance Normalization operation. Then, the scene motion $\bm{m}^{s}$ is processed through three convolutional layers to obtain the motion-adapted standard deviation $\bm{\gamma}^i \in \mathbb{R}^{C \times H \times W}$ and the mean $\bm{\beta}^i \in \mathbb{R}^{C \times H \times W}$. Here, $C$, $W$, $H$ represent the channel, height, and width of the feature, respectively. It is important to note that both $\bm{\beta}^i$ and $\bm{\gamma}^i$ have spatial dimensions, enabling pixel-wise guidance of the entire scene motion. This process is formulated as:
\begin{equation} \label{Eq.2}
\bm{f}^{i'} = \bm{\gamma}^{i}_{C,H,W} \left( \bm{m}^{s} \right) \frac{\bm{f}^{i}_{C,H,W} - \bm{\mu}^{i}_{C}}{\bm{\sigma}^{i}_{C}} + \bm{\beta}^{i}_{C,H,W}  \left( \bm{m}^{s} \right),
\end{equation}
where $\bm{\mu}^{i}_{C}$ and $\bm{\sigma}^{i}_{C}$ are the mean and standard deviation of $\bm{f}^i$ along the channel dimension.

With the proposed Mixed Motion Modeling and Mixed-Control Diffusion, along with the incorporation of the MAN module after each down-sampling block in the Reference UNet, we outline the complete pipeline of our MikuDance. Furthermore, to enhance MikuDance's ability to accommodate various styles of character art and the dynamics of large-scale camera movements, we present a mixed-source training approach in the next section.

\begin{figure}[t]
\vspace{-.5cm}
\setlength{\abovecaptionskip}{-.2cm}
\setlength{\belowcaptionskip}{-.5cm}
\begin{center}
   \includegraphics[width=1.0\linewidth]{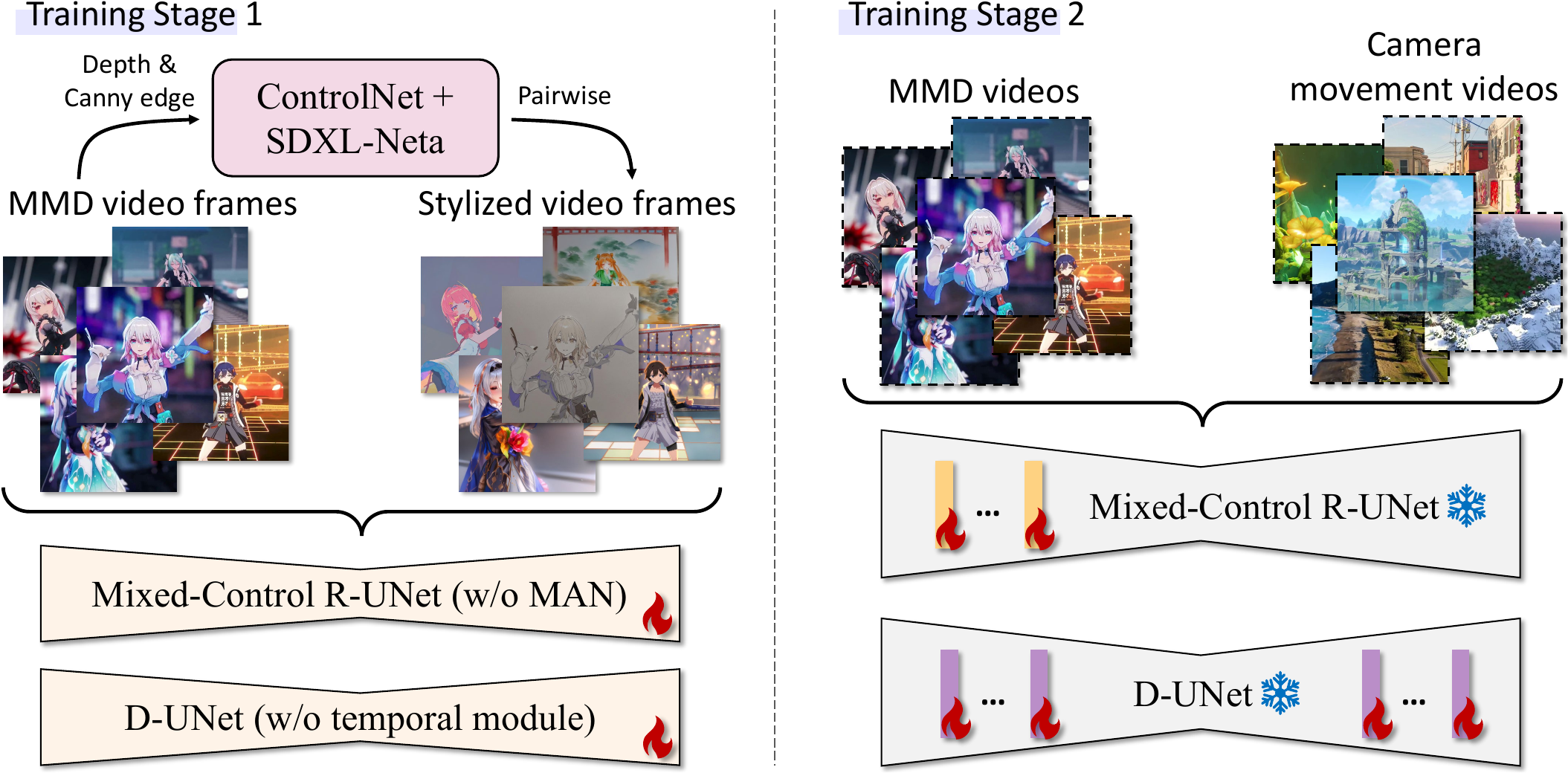}
\end{center}
   \caption{\textbf{The mixed-source training approach.} We utilize synthetic stylized video frames and non-character videos in the two training stages, respectively, to enhance generalizability.}
\label{fig:4}
\end{figure}

\subsection{Mixed-Source Training Approach} \label{sec:train}
Considering that image animation is a data-intensive task, proposing an effective data and training pipeline is as crucial as the model itself. In our MikuDance, as illustrated in Figure \ref{fig:4}, we adopt a mixed-source training approach with two-stage training stages.

In the first stage, training is conducted using pair-wise video frames, without incorporating the MAN module of the Reference UNet or the temporal module of the Denoising UNet. Different from existing methods \cite{hu2024animate, wang2024unianimate}, we randomly mix stylized pair-wise frames by concatenating the initial frames along the spatial dimension and utilize the depth and edge-controlled anime SDXL model \cite{podellsdxl, zhang2023adding}, known as SDXL-Neta \cite{neta}, to transfer the art style while preserving the image content. Additionally, to simulate the inference process in which the reference character art is irrelevant to the driving pose, we randomly select reference frames that are not involved in the target sequence.

In the second stage, both the MAN module and the temporal module are incorporated into our Mixed-Control Diffusion model, while the other parameters remain frozen during this phase. The training data in this stage consists of mixed MMD video clips and camera movement videos that do not include characters. Importantly, we randomly drop the pose and motion guidance during the two-stage training to enhance the robustness of our MikuDance.
\section{Experiments}
\noindent\textbf{Datasets.}
To train our MikuDance, we collected an MMD video dataset comprising 3,600 animations created by artists, all rendered from 3D models. We split these videos into approximately 120,000 clips, which together include over 10.2 million frames. Additionally, we incorporated around 3,500 non-character camera movement videos in the second training stage. For quantitative evaluation, we used 100 MMD videos that were not included in the training set, with their first frames serving as reference images. We used Xpose \cite{yang2023Xpose} for character pose and DROID-SLAM \cite{teed2021droid} for camera pose extraction. For qualitative evaluation, all character art was randomly generated using SDXL-Neta \cite{neta} and the driving videos were unseen during training.

\vspace{1mm}
\noindent\textbf{Implementation details.}
We implement MikuDance using the SD-1.5 framework \cite{rombach2022high} and PyTorch \cite{paszke2019pytorch}. Experiments are conducted on 16 NVIDIA A800 GPUs. In the first training stage, the video frames are center-cropped and resized to a resolution of $768 \times 768$. Training is conducted for 120,000 steps with a batch size of 128. In the second training stage, we train the MAN module and the temporal module for 60,000 steps using 24-frame video sequences and a batch size of 16. Both learning rates are set to 1e-5, and the dropout ratio for the pose and scene motion guidance is set to 0.2. During inference, we use a DDIM sampler for 20 denoising steps. We adopt the temporal aggregation method described in \cite{tseng2023edge} to generate long videos.

\vspace{1mm}
\noindent\textbf{Evaluation metrics.}
Following DISCO \cite{wang2024disco}, we evaluate the results from two aspects: image and video. To assess image quality, we report frame-wise FID \cite{heusel2017gans}, SSIM \cite{wang2004image}, LISPIS \cite{zhang2018unreasonable}, PSNR \cite{hore2010image}, and L1. For video quality, we concatenate every consecutive 16 frames to form a sample, from which we report FID-VID \cite{balaji2019conditional} and FVD \cite{unterthiner2018towards}.

\subsection{Qualitative Results}

\begin{figure*}[t]
\vspace{-.8cm}
\setlength{\abovecaptionskip}{-.3cm}
\setlength{\belowcaptionskip}{-.2cm}
\begin{center}
   \includegraphics[width=1.0\linewidth]{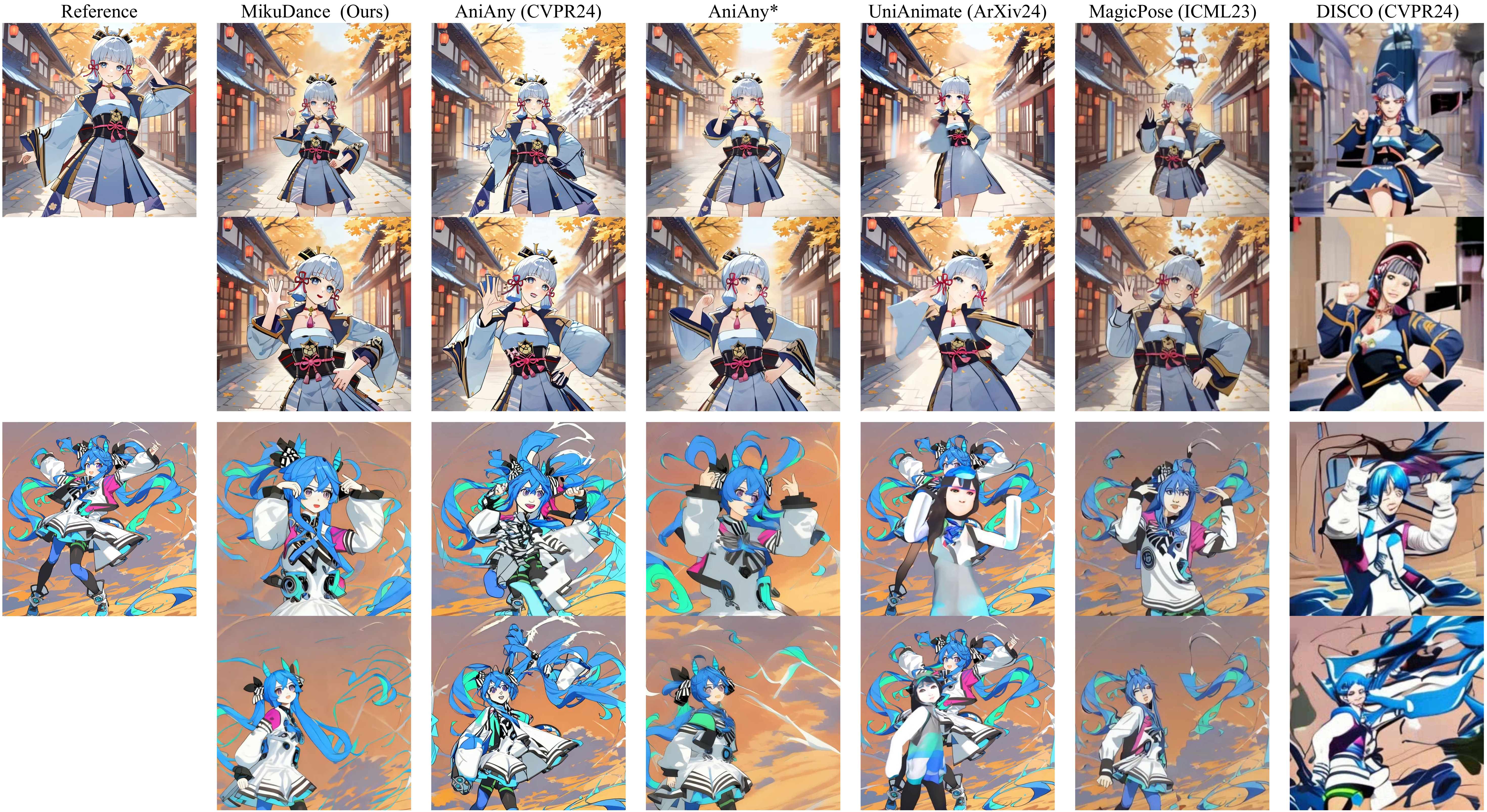}
\end{center}
   \caption{\textbf{Comparison with the baselines.} AniAny* is the fine-tuned version of the AniAny model, trained on our MMD video dataset.}
\label{fig:exp_sota}
% \end{figure*}

% \begin{figure*}[t]
% \vspace{-.8cm}
\setlength{\abovecaptionskip}{-.3cm}
\setlength{\belowcaptionskip}{-.3cm}
\begin{center}
   \includegraphics[width=1.0\linewidth]{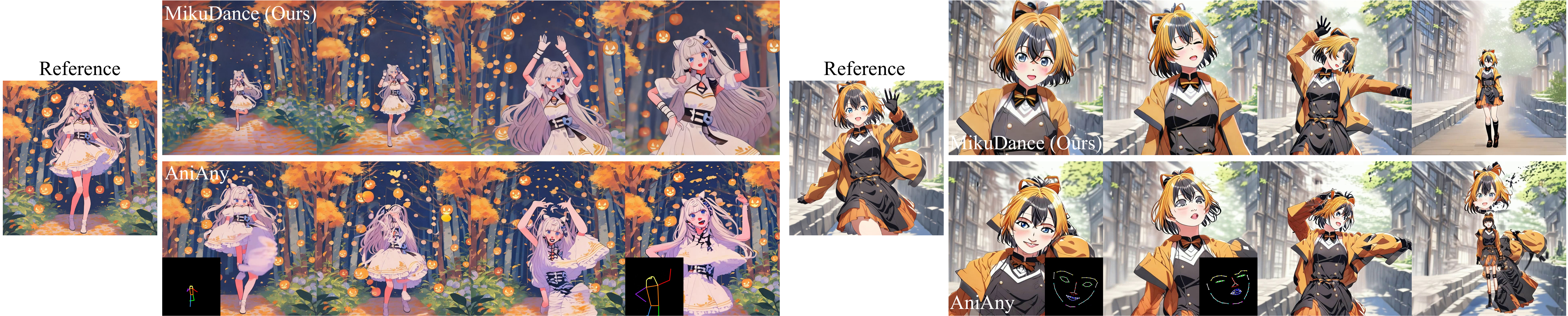}
\end{center}
   \caption{\textbf{Experiments on high-dynamic motion guidance.} Large camera movements (left) and significant pose variations (right).}
\label{fig:exp_motion}
\end{figure*}

\noindent\textbf{Comparison with the baselines.} We compare our MikuDance with recent human video generation methods, including Animate Anyone (AniAny) \cite{hu2024animate}, DISCO \cite{wang2024disco}, MagicPose \cite{chang2023magicpose}, and UniAnimate \cite{wang2024unianimate}, all of which claim the capability to animate anime-style characters in their official reports. Additionally, we implemented AniAny* by fine-tuning the model on our MMD video dataset.

The results in Figure \ref{fig:exp_sota} show that AniAny, MagicPose, and UniAnimate fail to address the misalignment of character shape and scale, resulting in character distortion in their outputs. While DISCO uses independent ControlNets to process background and foreground features, its results suffer from scene collapse when animating character art. Although AniAny* is specifically fine-tuned on the anime-style dataset, its results show limited improvements and blurring in high-dynamic motion, as the pipeline fails to account for background scene motion. Notably, MikuDance effectively handles complex reference and motion guidance, delivering high-quality and vivid animation results.

\begin{figure*}[t]
\vspace{-.8cm}
\setlength{\abovecaptionskip}{-.3cm}
\setlength{\belowcaptionskip}{-.3cm}
\begin{center}
   \includegraphics[width=1.0\linewidth]{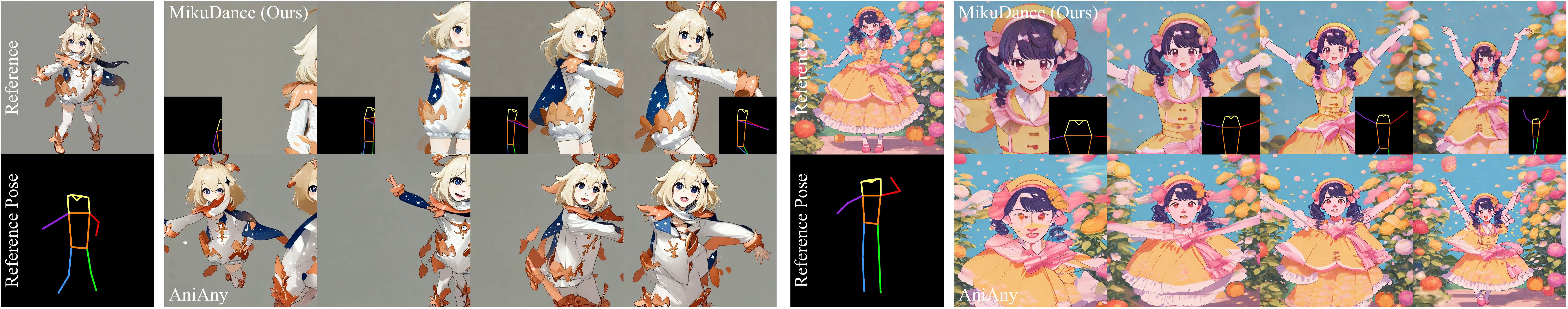}
\end{center}
   \caption{\textbf{Experiments on misalignments between reference images and driving poses.} Spatial (left) and scale (right) misalignments.}
\label{fig:exp_mis}
% \end{figure*}

% \begin{figure*}[t]
% \vspace{-.4cm}
\setlength{\abovecaptionskip}{-.3cm}
\setlength{\belowcaptionskip}{-.2cm}
\begin{center}
   \includegraphics[width=1.0\linewidth]{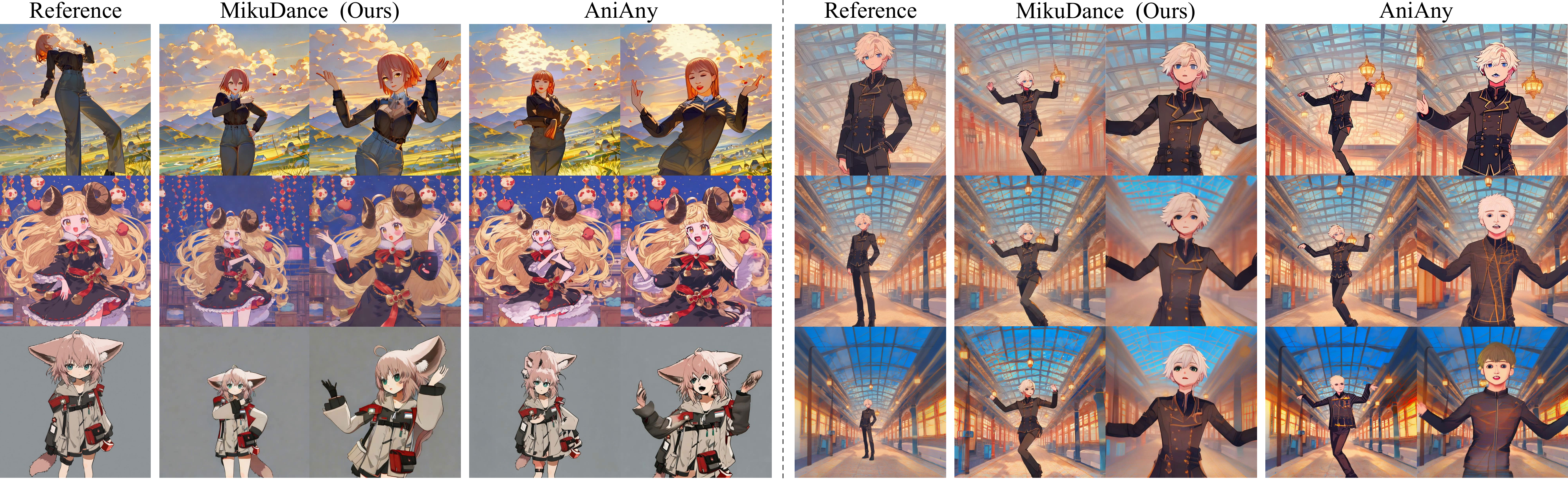}
\end{center}
   \caption{\textbf{Experiments on various shapes (the left part) and scales (the right part) of the reference character art.}}
\label{fig:exp_shape_scale}
% \end{figure*}

% \begin{figure*}[t]
% \vspace{-1cm}
\setlength{\abovecaptionskip}{-.3cm}
\setlength{\belowcaptionskip}{-.3cm}
\begin{center}
   \includegraphics[width=1.0\linewidth]{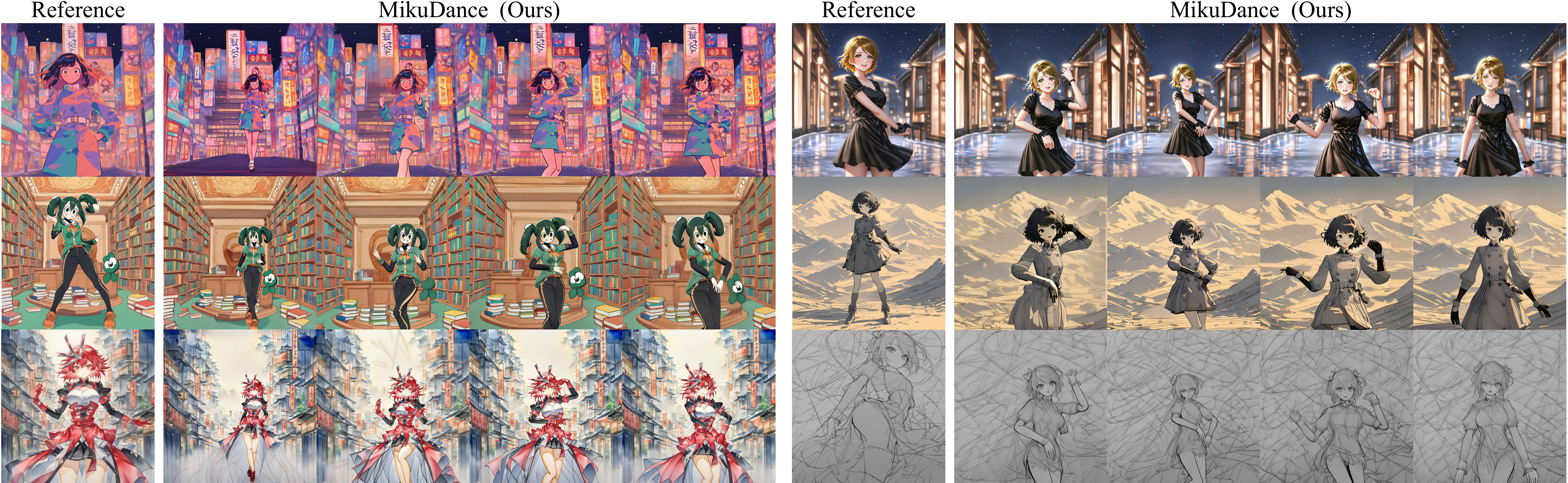}
\end{center}
   \caption{\textbf{Experiments on various styles of the reference character art.} Please see Appendix for more results.}
\label{fig:exp_style}
\end{figure*}

\vspace{1mm}
\noindent\textbf{High-dynamic motion.} A key highlight of MikuDance is its ability to handle high-dynamic motion guidance in animation, which goes beyond the simple actions used in existing methods. As shown in Figure \ref{fig:exp_motion}, the character performs large dance movements with a fast-moving camera. Despite these challenging conditions, MikuDance, equipped with our Mixed Motion Modeling approach, demonstrates remarkable robustness, delivering high-fidelity animation results that effectively capture the dramatic visual impact.

\vspace{1mm}
\noindent\textbf{Reference-guidance misalignments.} Another key contribution of MikuDance is its implicit alignment of the reference character with motion guidance. As illustrated in Figure \ref{fig:exp_mis}, two examples show significant spatial and scale misalignments between the guidance and the reference. In such cases, existing methods like AniAny struggle to animate the reference character effectively, whereas MikuDance successfully manages these complexities and generates coherent animations.

% \vspace{1mm}
\noindent\textbf{Various shapes and scales.} MikuDance effectively handles variations in character shapes and scales. As shown in the left part of Figure \ref{fig:exp_shape_scale}, characters with distinct body shapes, various poses, and different clothing are precisely driven by the same motion guidance. In the right part of Figure \ref{fig:exp_shape_scale}, MikuDance demonstrates its ability to implicitly align characters of varying scales, preserving each character’s unique features and producing reasonable animation results

\vspace{1mm}
\noindent\textbf{Generalizability on various art styles.} As illustrated in Figure \ref{fig:exp_style}, MikuDance, leveraging our mixed-source training approach, can handle a wide range of art styles, including but not limited to celluloid, antiquity, and line sketch. This high level of generalizability opens up broad prospects for real-world applications.

\begin{figure*}[t]
\vspace{-.8cm}
\setlength{\abovecaptionskip}{-.3cm}
\setlength{\belowcaptionskip}{-.1cm}
\begin{center}
   \includegraphics[width=1.0\linewidth]{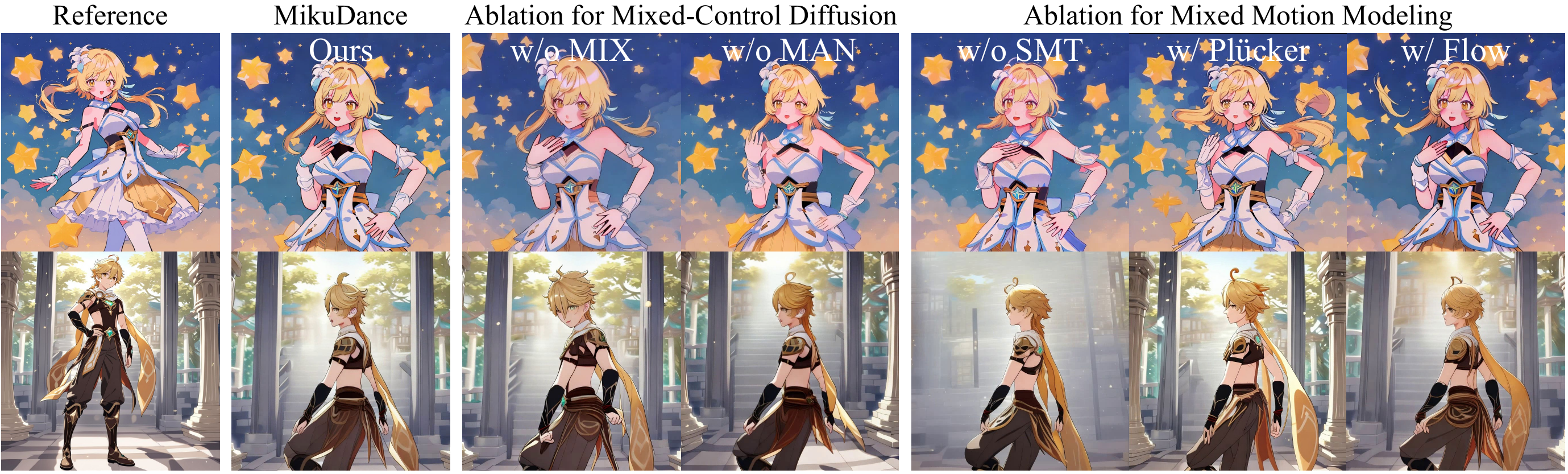}
\end{center}
   \caption{\textbf{Ablation experiments on the key designs of MikuDance.} MIX, SMT, MAN, Plücker, and Flow are defined in Section \ref{ablsty}.}
\label{fig:exp_abl}
\end{figure*}

\vspace{1mm}
\noindent\textbf{Ablation study.} We conduct ablation experiments to verify the key designs of our MikuDance, as shown in Figure \ref{fig:exp_abl}, which include the mixed-control architecture (MIX), the MAN module, and the SMT strategy. \label{ablsty}

To evaluate the mixed-control design, we implemented a pipeline (w/o MIX) inspired by AniAny, utilizing an independent Reference UNet to process the reference image and two ControlNets to separately adapt the character and scene motion guidance. The results indicate that this conventional pipeline fails to account for scale differences between the character art and the driving guidance, leading to a mismatched appearance of the character's face and pose.

To evaluate the effectiveness of the MAN module, we implemented a pipeline without MAN (w/o MAN) that simply concatenates scene motion with character motion and inputs them together into the Reference UNet. While this approach yields better results than a pipeline without scene motion guidance (w/o SMT), it remains inferior to the results achieved by MikuDance. This is because the MAN module injects global motion through spatial-aware normalization, effectively complementing the local motion.

To evaluate the SMT strategy, we conducted three experiments: one pipeline without incorporating scene motion (w/o SMT), and two pipelines that replaced the scene motion with Plücker embedding (w/ Plücker) and optical flow (w/ Flow), respectively. However, the results from these alternative approaches were inferior to our SMT strategy, showing noticeable artifacts and inconsistencies in the dynamic backgrounds. The pixel-wise scene motion extracted by SMT proved to be a more effective representation for guiding background motion due to its domain consistency with the character motion guidance.

Compared to the ablative studies introduced above, our MikuDance effectively addresses the misalignment and high-dynamic challenges in animating character art.

\subsection{Quantitative Results}
Table \ref{tab:1} presents quantitative comparisons between MikuDance and the baseline methods. It is important to note that the metrics reported in our paper are lower than those in previous studies, as the entire scene in our testing videos is highly dynamic, unlike the static backgrounds used in earlier datasets. Nevertheless, the results demonstrate that MikuDance achieves state-of-the-art performance across all image and video metrics. Additionally, the ablation results confirm the effectiveness of the key design elements in MikuDance. In summary, by incorporating the proposed mixed motion dynamics techniques, MikuDance can animate a wide range of characters and generate high-quality image and video results.

\begin{SCtable*}[]
% \setlength{\abovecaptionskip}{-.01cm}
% \setlength{\belowcaptionskip}{-.2cm}
% \vspace{-.5cm}
\small{
\caption{ \textbf{Quantitative comparisons with baselines and ablative experiments.} AniAny* is the fine-tuned version of the AniAny model. MIX, SMT, MAN, Plücker, and Flow are defined in Section \ref{ablsty}. The best results are highlighted in bold, and the second-best are underlined. MikuDance achieves superior results across all metrics.}%
\label{tab:1}
\begin{tabular}{ l | c c c c c | c c } 
\toprule[1.5pt]
  \multirow{2}{*}{\textbf{Methods}} & \multicolumn{5}{c|}{\textbf{Image}} & \multicolumn{2}{c}{\textbf{Video}} \\
  \cmidrule{2-8}
		& \textbf{FID}$_{\downarrow}$ & \textbf{SSIM}$_{\uparrow}$ & \textbf{PSNR}$_{\uparrow}$ & \textbf{LISPIS}$_{\downarrow}$ & \textbf{L1}$_{\downarrow}$ & \textbf{FID-VID}$_{\downarrow}$ &  \textbf{FVD}$_{\downarrow}$  \\
            \midrule[0.5pt]
            AniAny \cite{hu2024animate} & 43.945 & 0.488 & 12.530 & 0.548 & 7.307E-05 & 38.179 & 846.414 \\
            AniAny* & 28.833 & 0.526 & 13.610 & 0.517 & 6.229E-05 & 26.764 & 575.304 \\
            DISCO \cite{wang2024disco} & 59.221 & 0.313 & 10.732 & 0.615 & 9.248E-05 & 46.852 & 923.921 \\
            MagicPose \cite{chang2023magicpose} & 44.258 & 0.424 & 12.357 & 0.554 & 7.767E-05 & 41.347 & 886.691 \\
            UniAnimate \cite{wang2024unianimate} & 47.328 & 0.417 & 12.074 & 0.571 & 7.930E-05 & 40.924 & 882.245 \\
            \midrule[0.5pt]
            MikuDance w/o MIX  & 27.315 & 0.523 & 14.004 & 0.528 & 5.860E-05 & 24.124 & 541.453  \\
            MikuDance w/o MAN  & \underline{24.985} & \underline{0.542} & \underline{14.501} & \underline{0.505} & \underline{5.753E-05} & 23.366 & 509.342  \\
             MikuDance w/o SMT  & 25.472 & 0.534 & 14.312 & 0.512 & 5.911E-05 & 23.362 & 517.673  \\
             MikuDance w/ Plüc.  & 25.918 & 0.538 & 14.261 & 0.510 & 6.011E-05 & 23.471 & 521.853  \\
             MikuDance w/ Flow & 26.141 & 0.516 & 14.088 & 0.523 & 5.925E-05 & \underline{23.079} & \underline{505.533}  \\
            MikuDance (Ours) & \textbf{24.597} & \textbf{0.576} & \textbf{14.592} & \textbf{0.493} & \textbf{5.726E-05} & \textbf{22.868} & \textbf{502.380}  \\
		\bottomrule[1.5pt]
\end{tabular}}
\end{SCtable*}

\begin{figure}[t]
\vspace{-.2cm}
\setlength{\abovecaptionskip}{-.2cm}
\setlength{\belowcaptionskip}{-.5cm}
\begin{center}
   \includegraphics[width=1.0\linewidth]{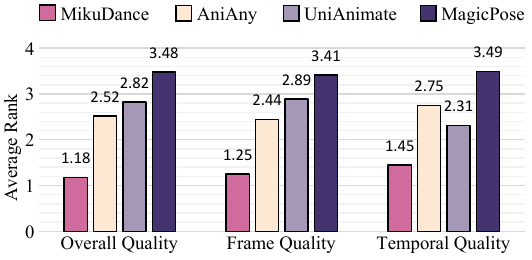}
\end{center}
   \caption{\textbf{User Study.} The smaller value means the better quality.}
\label{fig:userstudy}
\end{figure}

\vspace{1mm}
\noindent\textbf{User study.} We invited 50 volunteers and gave them 20 videos to evaluate the performance of our MikuDance against the baseline methods. Each video includes one motion guidance and four anonymous animation results. We ask users to rank the four results in overall quality, frame quality, and temporal quality. After excluding abnormal questionnaires, the average rank of the methods is summarized in Figure \ref{fig:userstudy}. Our MikuDance outperforms the baseline methods by a large margin and more than 97\% of users prefer the animation generated by our MikuDance.
\section{Conclusions}
In this work, we propose MikuDance, a new animation pipeline designed to generate high-dynamic animations for in-the-wild character art. MikuDance incorporates two key techniques: Mixed Motion Modeling and Mixed-Control Diffusion. Mixed Motion Modeling enables the representation of large-scale character and scene motions within a unified reference space, while Mixed-Control Diffusion addresses misalignment between characters and motion guidance. To support diverse art styles, we also employ a mixed-source training approach to enhance generalizability. Extensive experiments demonstrate that MikuDance achieves state-of-the-art performance compared to baseline methods.

% \vspace{1mm}
\noindent\textbf{Limitations.} We acknowledge that some generated animations exhibit background distortions and artifacts. This issue stems from the 3D-agnostic challenge in image animation, making scene reconstruction in dynamic cameras an ill-posed problem that requires further investigation.

{
    \small
    \bibliographystyle{ieeenat_fullname}
    \bibliography{main}
}


\clearpage
%% 页数从1开始
\setcounter{page}{1} 
%% 重新设置节编号 
\renewcommand\thesection{\Alph{section}}
%% 重新计数
\setcounter{section}{0}
% \setcounter{figure}{0}
% \setcounter{table}{0}
%%%%%%%%%%%%%%%%%%%%%%%%%%%%%%
% \numberwithin{equation}{section}
% \numberwithin{figure}{section}
% \numberwithin{table}{section}

\maketitlesupplementary

% \begin{SCtable*}[]
% % \setlength{\abovecaptionskip}{-.01cm}
% % \setlength{\belowcaptionskip}{-.2cm}
% % \vspace{-.5cm}
% \small{
% \caption{ \textbf{Quantitative comparisons on foreground-only (FO) results.} AniAny* is the fine-tuned version of the AniAny model. The best results are highlighted in bold, and the second-best are underlined.}%
% \label{tab:1}
% \begin{tabular}{ l | c c c c c | c c } 
% \toprule[1.5pt]
%   \textbf{\multirow{2}{*}\textbf{{Methods}}} & \multicolumn{5}{c|}{\textbf{Image}} & \multicolumn{2}{c}{\textbf{Video}} \\
%   \cmidrule{2-8}
% 		& \textbf{FID}$_{\downarrow}$ & \textbf{SSIM}$_{\uparrow}$ & \textbf{PSNR}$_{\uparrow}$ & \textbf{LISPIS}$_{\downarrow}$ & \textbf{L1}$_{\downarrow}$ & \textbf{FID-VID}$_{\downarrow}$ &  \textbf{FVD}$_{\downarrow}$  \\
%             \midrule[0.5pt]
%             AniAny & 43.945 & 0.488 & 12.530 & 0.548 & 7.307E-05 & 38.179 & 846.414 \\
%             AniAny* & 28.833 & 0.526 & 13.610 & 0.517 & 6.229E-05 & 26.764 & 575.304 \\
%             DISCO & 59.221 & 0.313 & 10.732 & 0.615 & 9.248E-05 & 46.852 & 923.921 \\
%             MagicPose & 44.258 & 0.424 & 12.357 & 0.554 & 7.767E-05 & 41.347 & 886.691 \\
%             UniAnimate & 47.328 & 0.417 & 12.074 & 0.571 & 7.930E-05 & 40.924 & 882.245 \\
%               MikuDance (Ours)  & \textbf{14.835} & \textbf{0.846} & \textbf{17.809} & \textbf{0.137} & \textbf{2.060E-05} & \textbf{6.765} & \textbf{194.124}  \\
% 		\bottomrule[1.5pt]
% \end{tabular}}
% \end{SCtable*}

\begin{table*}[b]
\setlength{\belowcaptionskip}{-.4cm}
% \vspace{-.5cm}
\centering{
\begin{tabular}{ l | c c c c c | c c } 
\toprule[1.5pt]
 \multirow{2}{*}{\textbf{Methods}} & \multicolumn{5}{c|}{\textbf{Image}} & \multicolumn{2}{c}{\textbf{Video}} \\
  \cmidrule{2-8}
		& \textbf{FID}$_{\downarrow}$ & \textbf{SSIM}$_{\uparrow}$ & \textbf{PSNR}$_{\uparrow}$ & \textbf{LISPIS}$_{\downarrow}$ & \textbf{L1}$_{\downarrow}$ & \textbf{FID-VID}$_{\downarrow}$ &  \textbf{FVD}$_{\downarrow}$  \\
            \midrule[0.5pt]
            AniAny & 27.927 & 0.625 & 14.831 & 0.431 & 3.307E-05 & 15.197 & 326.842 \\
            AniAny* & 17.037 & 0.801 & 17.194 & 0.202 & 2.284E-05 & 7.546 & 211.384 \\
            DISCO & 31.221 & 0.533 & 13.155 & 0.518 & 4.824E-05 & 22.828 & 564.892 \\
            MagicPose & 25.248 & 0.695 & 15.240 & 0.362 & 3.014E-05 & 13.248 & 299.912 \\
            UniAnimate & 29.818 & 0.601 & 14.147 & 0.448 & 3.902E-05 & 18.416 & 381.485 \\
              MikuDance (Ours)  & \textbf{14.835} & \textbf{0.846} & \textbf{17.809} & \textbf{0.137} & \textbf{2.060E-05} & \textbf{6.765} & \textbf{194.124}  \\
		\bottomrule[1.5pt]
\end{tabular}}
\caption{ \textbf{Quantitative comparisons on foreground-only results.}}%
\label{tab:fo}
% \vspace{-.8cm}
\end{table*}

\section{Supplementary on Experiments}
\label{sec:sup_exp}

% \vspace{1mm}
\noindent\textbf{Quantitative comparisons on foreground-only results.} The quantitative results of previous methods are primarily evaluated on videos with static backgrounds and simple human motions. In contrast, our test dataset includes high-dynamic motions in both the foreground and background. To provide a clearer evaluation, we report additional comparisons on foreground-only results in Table \ref{tab:fo}, which remove complex backgrounds using a segmentation mask generated by BiRefNet (CAAI-AIR'24) and focus solely on the quality of character animation. The results show that MagicPose outperforms AniAny in character quality, as it better preserves the character's appearance. Among all methods, our MikuDance consistently achieves superior results across all metrics.

\begin{figure}[b]
% \vspace{-.5cm}
\setlength{\abovecaptionskip}{-.2cm}
\begin{center}
   \includegraphics[width=1.0\linewidth]{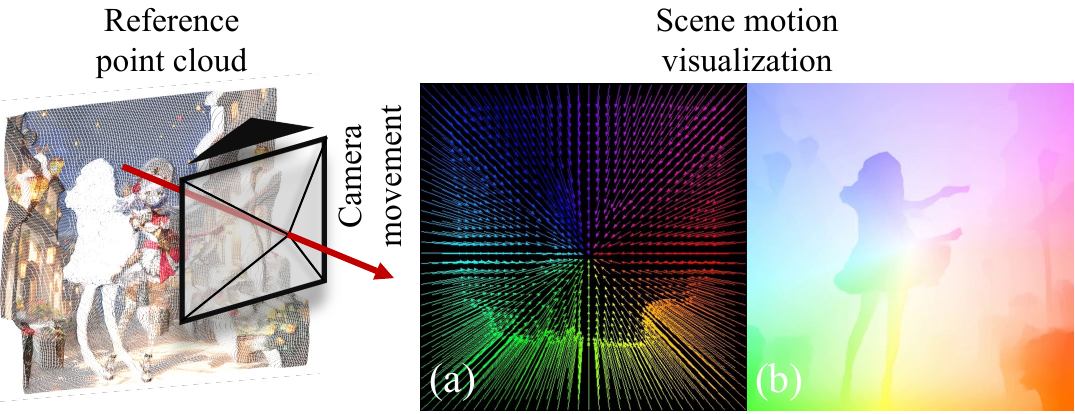}
\end{center}
   \caption{\textbf{Visualization of the scene motion.}}
\label{fig:sup_vis_mo}
\end{figure}

\vspace{1mm}
\noindent\textbf{Scene motion visualization.} Figure \ref{fig:sup_vis_mo} presents a visualized example of scene motion, depicted using directed line segments (a) and dense flow (b).

\begin{figure*}[t]
\vspace{-.2cm}
\setlength{\abovecaptionskip}{-.3cm}
\setlength{\belowcaptionskip}{-.3cm}
\begin{center}
   \includegraphics[width=1.0\linewidth]{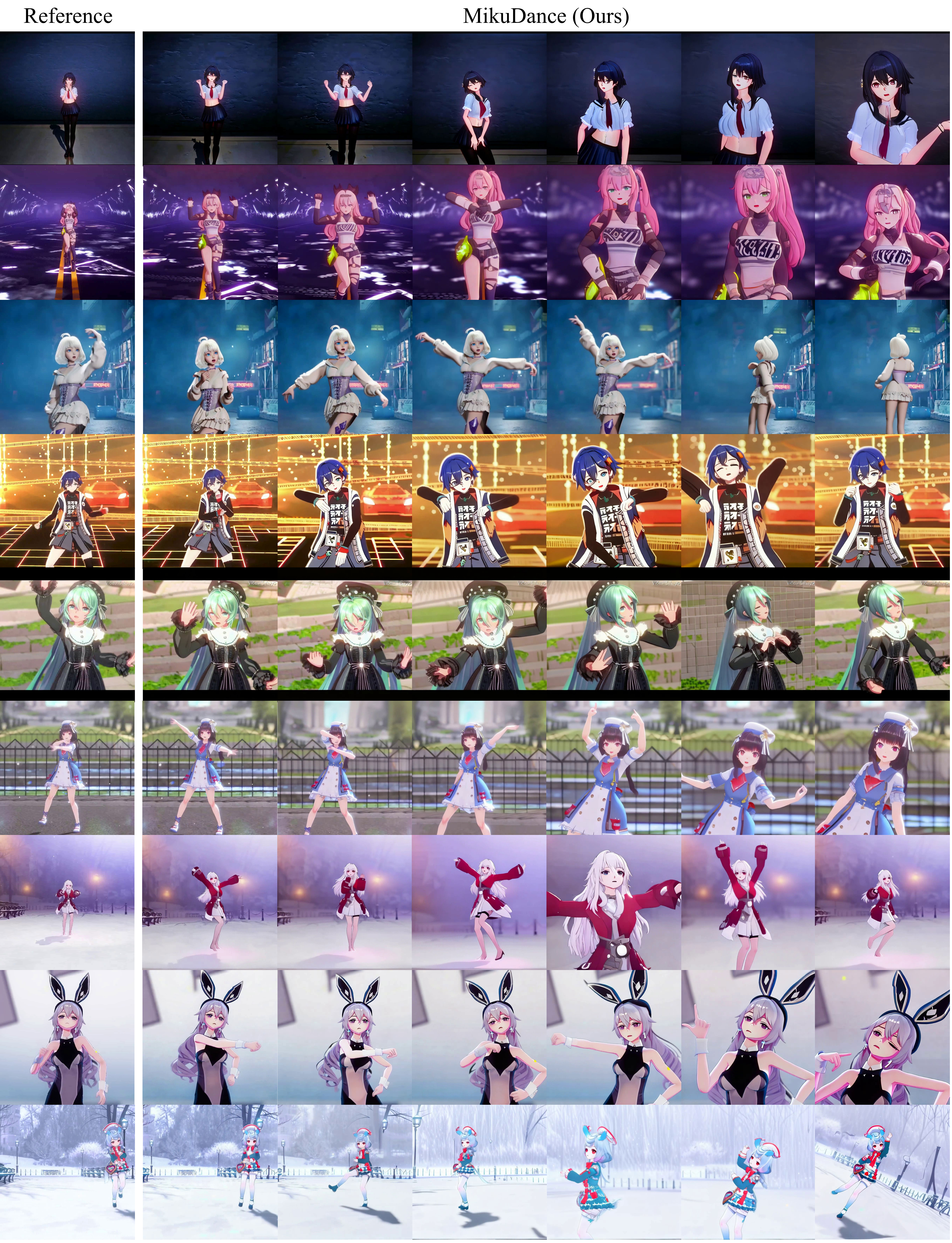}
\end{center}
   \caption{\textbf{Supplementary experiments on animating video frames.}}
\label{fig:sup_frames}
\end{figure*}

\vspace{1mm}
\noindent\textbf{Animating video frames.} Since quantitative evaluation requires ground-truth video to calculate metrics, we collected an unseen test video dataset and used the first frames as reference images. Additionally, most results reported in prior works, such as Animate Anyone and DISCO, involve the animation of video frames where the reference image is naturally aligned with the motion guidance. Examples of video frame animations generated by MikuDance are illustrated in Figure \ref{fig:sup_frames}. These results show that, when handling the simpler task of animating video frames, MikuDance exhibits stable, high-quality performance even with complex and high-dynamic motion guidance. Notably, the primary motivation behind MikuDance is to animate in-the-wild character art, offering broader application potential compared to prior works focused on animating video frames.

\begin{figure*}[t]
\vspace{-.2cm}
\setlength{\abovecaptionskip}{-.3cm}
\setlength{\belowcaptionskip}{-.3cm}
\begin{center}
   \includegraphics[width=1.0\linewidth]{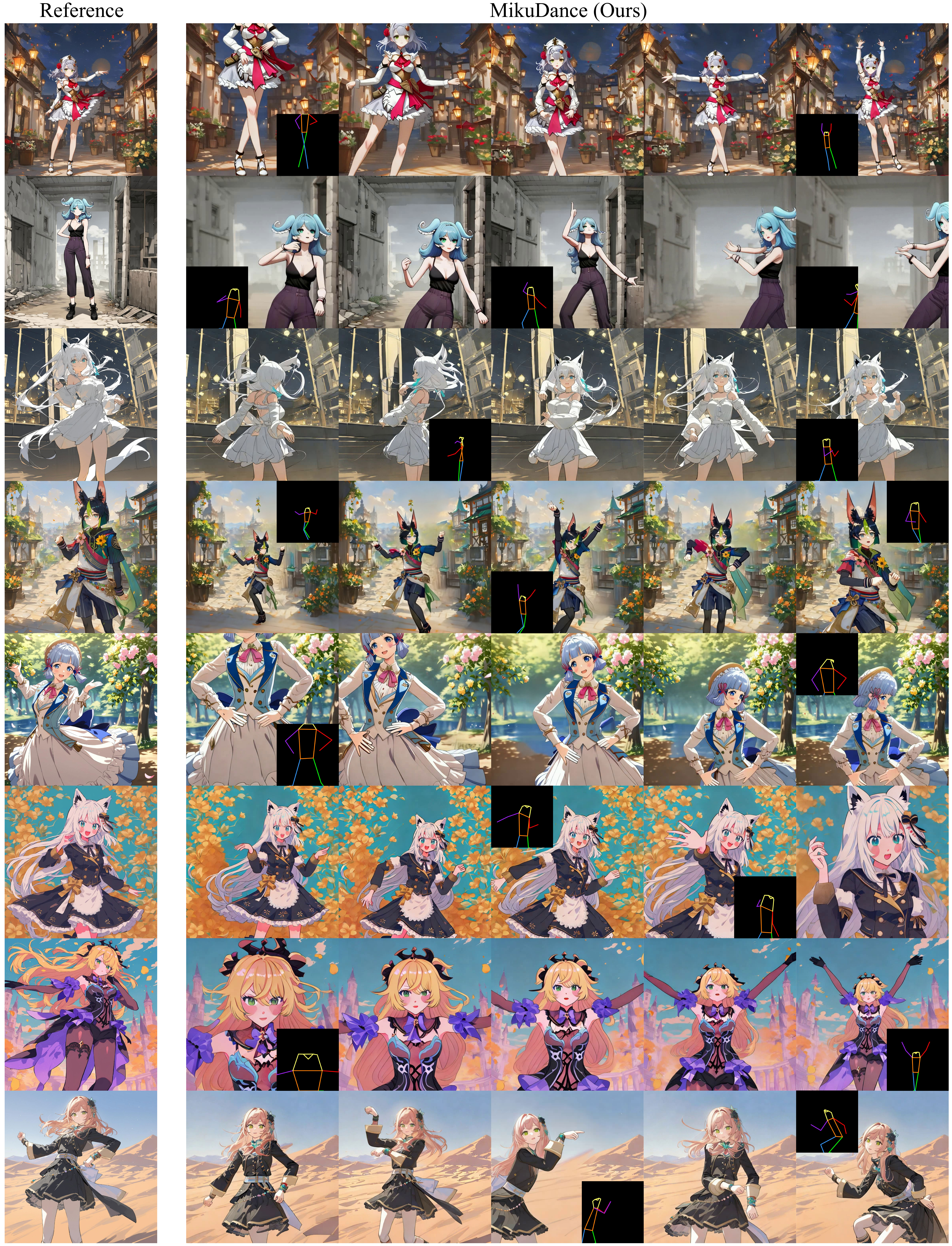}
\end{center}
   \caption{\textbf{Additional results on high-dynamic motions.}}
\label{fig:sup_motion}
\end{figure*}

\vspace{1mm}
\noindent\textbf{Additional results on high-dynamic motions.} In Figure \ref{fig:sup_motion}, we present additional animation results generated by MikuDance under high-dynamic motion guidance, which includes large pose variations, extensive camera movements, turning around, and incomplete body postures. These results further demonstrate MikuDance's robust ability to handle complex character and scene motion.

\begin{figure*}[t]
\vspace{-.2cm}
\setlength{\abovecaptionskip}{-.3cm}
\setlength{\belowcaptionskip}{-.3cm}
\begin{center}
   \includegraphics[width=1.0\linewidth]{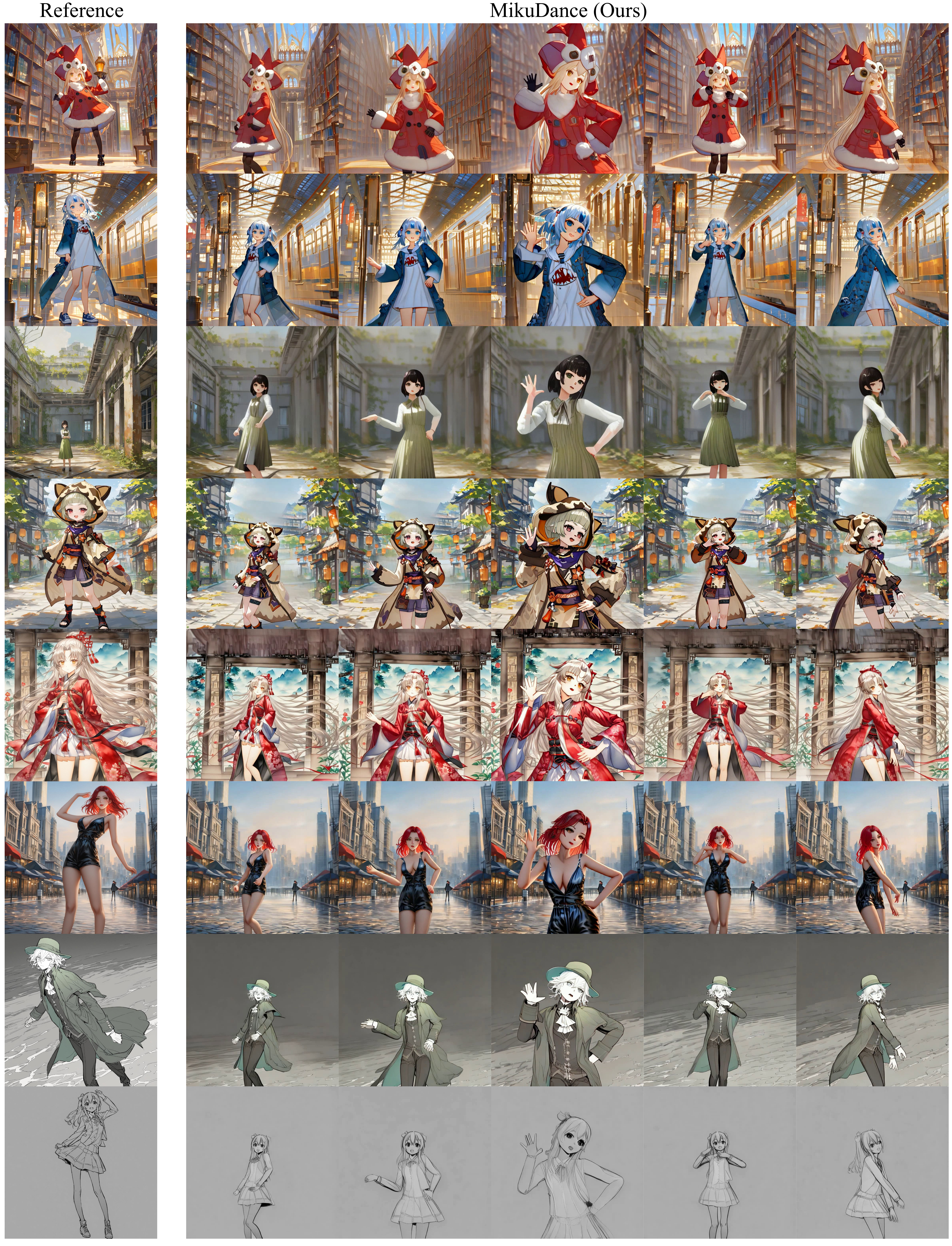}
\end{center}
   \caption{\textbf{Additional results on various shapes and styles.}}
\label{fig:sup_ss}
\end{figure*}

\vspace{1mm}
\noindent\textbf{Additional results on various shapes and styles.} MikuDance demonstrates strong generalizability across a wide range of characters and art styles. Additional results are illustrated in Figure \ref{fig:sup_ss}. Remarkably, even in extremely challenging cases, such as the one shown in the third row of Figure \ref{fig:sup_ss}, MikuDance consistently generates vivid animations that preserve both the character's appearance and the scene structure.

\section{Demo Videos and User Study Details}
\label{sec:demo}
We provide two demo videos. MikuDance-Results.mp4 showcases the animation results presented in our paper. MikuDance-Queencard.mp4 is a complete Music Video (MV) with animations fully generated by MikuDance, using reference character art created by SDXL-Neta.

\begin{figure*}[t]
% \vspace{-.2cm}
\setlength{\abovecaptionskip}{-.3cm}
\setlength{\belowcaptionskip}{-.3cm}
\begin{center}
   \includegraphics[width=1.0\linewidth]{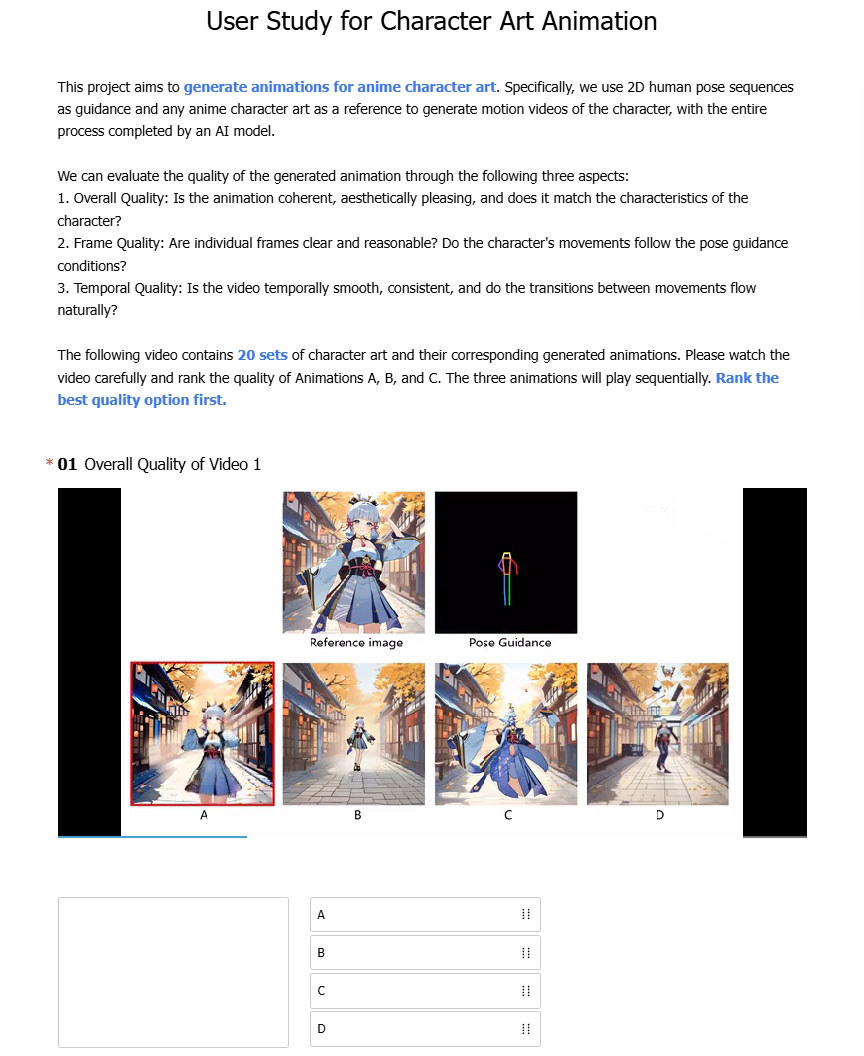}
\end{center}
   \caption{\textbf{An example of our user study webpage.}}
\label{fig:sup_user}
\end{figure*}

Figure \ref{fig:sup_user} shows an example of the user study webpage. In one case, 50 volunteers were invited to rank four anonymous results based on three carefully defined evaluation criteria. We provided 20 videos for the volunteers to evaluate, and the average time required to complete the entire questionnaire was over 30 minutes.

% \small{

% \bibitem{zheng2024birefnet}Zheng, P., Gao, D., Fan, D., Liu, L., Laaksonen, J., Ouyang, W. \& Sebe, N. Bilateral Reference for High-Resolution Dichotomous Image Segmentation. {\em CAAI Artificial Intelligence Research}. \textbf{3} pp. 9150038 (2024)

% \end{thebibliography}
% }

\end{document}